\documentclass[a4paper]{svproc}
\usepackage{multicol}
\usepackage[bookmarks=true]{hyperref}
\usepackage{amsfonts}
\usepackage{enumerate}
\usepackage{wrapfig}
\usepackage{graphics} 
\usepackage{epsfig} 
\usepackage{amsmath} 
\usepackage{amssymb}  
\usepackage{array}
\usepackage{subcaption}
\usepackage[font=footnotesize]{caption}
\usepackage{leftidx}
\usepackage{hyperref}
\usepackage[tight]{units}
\usepackage{color}
\renewcommand{\theequation}{CT\arabic{equation}}
\usepackage[sectionbib,numbers]{natbib}
\usepackage[subtle]{savetrees} 

\usepackage{url}

\newcommand{\myparagraph}[1]{\vspace{0.05in}\noindent\textbf{#1}}

\begin{document}
\mainmatter              
\title{Certified Grasping}
\titlerunning{Certified Grasping}  
%
\author{Bernardo Aceituno-Cabezas, Jose Ballester, and Alberto Rodriguez}
\authorrunning{Aceituno-Cabezas et al.} 
%
\tocauthor{Bernardo Aceituno-Cabezas, Jose Ballester, and Alberto Rodriguez}
\institute{Massachusetts Institute of Technology, Cambridge MA, USA,\\
\email{\{aceituno,joseb,albertor\}@mit.edu}\\
}

\maketitle              

\begin{abstract}
This paper studies robustness in planar grasping from a geometric perspective. By treating grasping as a process that shapes the free-space of an object over time, we can define three types of certificates to guarantee success of a grasp: (a) invariance under an initial set, (b) convergence towards a goal grasp, and (c) observability over the final object pose. We develop convex-combinatorial models for each of these certificates, which can be expressed as simple semi-algebraic relations under mild-modeling assumptions. By leveraging these models to synthesize certificates, we optimize certifiable grasps of arbitrary planar objects composed as a union of convex polygons, using manipulators described as point-fingers. We validate this approach with simulations and real robot experiments, by grasping random polygons, comparing against other standard grasp planning algorithms, and performing sensorless grasps over different objects.
\end{abstract}

\section{Introduction}

The key question we study in this paper is that of robustness in the process of grasping an object. 
Can we ever certify that a planned grasp will work? \vspace{6pt}

The common approach to grasping is to plan an arrangement of contacts on the surface of an object. Experimental evidence shows an intuitive but also paradoxical observation: On one hand, most grasps do not work as expected since fingers do not deliver exactly the planned arrangement of contacts; on the other hand, many planned grasps still end up working and produce a stable hold of the object. 
These natural dynamics work within all grasping algorithms, often to their benefit, sometimes adversarially. 
\citet{mason2012autonomous} put it as: if we cannot put the fingers in the right place,  can we trust the fingers to \emph{fall where they may}?
In this paper we study the possibility to synthesize grasps for which the fingers have no other option than to do so.\vspace{6pt}

The notions of robustness and certification are central to the robotics community. However, formal approaches to synthesize robustness in grasping have been mostly limited to study the set of forces that a grasp can resist~\cite{bicchi2000robotic}, neglecting the key importance of the reaching motion towards that grasp.
%
Both the reaching motion and the end-grasp can encode robustness. In this paper we study the problem of synthesizing trajectories of a set of point fingers that converge onto an intended grasp of a polygonal planar object, naturally encoding robustness to uncertainty as part of the grasping process.
We start by proposing three different types of certificates that one can formulate at different stages of the grasping process:
%
%
\begin{itemize}
    \item \textbf{Invariance Certificate:} At the beginning of the grasping process, the object lies in an invariant set of its configuration space. In this paper we study the case when the object is geometrically trapped by fingers around it, i.e., the object is caged by the fingers~\cite{rodriguez2012caging}.
    \item \textbf{Convergence Certificate:} All configurations in the invariant set are driven towards a given end-grasp configuration. Intuitively, this is analogous to driving down the value of a scalar/energy function with only one minimum.
    \item \textbf{Observability Certificate:} The configuration of the object in the end-grasp is identifiable with the robot's contact or proprioceptive sensors after completing the grasp. In this work we characterize when the location of fingers is enough to recover the pose of the object, for which the condition is analogous to first-order form closure.
\end{itemize}

\begin{figure}[t]
    \centering
    \includegraphics{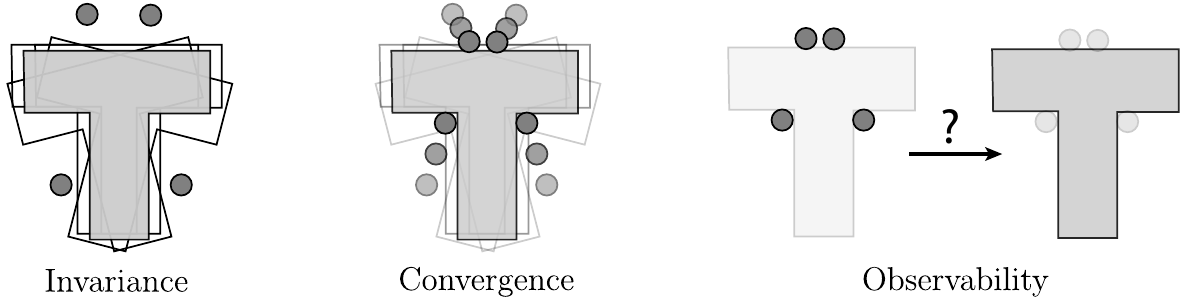}
    \caption{\textbf{Overview of grasping with certificates.} From a configuration space perspective, we say a grasp is certified to succeed when: 1) The robot bounds the object pose within an invariant set, and 2) the free-space converges to a single configuration. From this initial bound, we obtain an invariant set of configurations for which the grasp will always succeed. A third certificate, also valid for non-converging grasping processes, comes from requiring that the end-grasp configuration is observable.}
    \label{fig:my_label}
    \vspace{-12pt}
\end{figure}


Sections~\ref{sec:caging}, \ref{sec:convergence}, and \ref{sec:observability} derive a model for a particular formulation of each of these certificates. These models build on tools from convex-combinatorial optimization that decompose the configuration space of an object surrounded by fingers into free regions, and is based on recent work to formulate the caging synthesis problem as an optimization problem~\cite{aceituno-cabezas2019icra}. Section~\ref{sec:caging} summarizes the approach. 

The combination of the models for each of the three certificates yields a complete geometric model to synthesize grasping motions that reach certifiable grasps. Section~\ref{sec:results} describes the application of this model to robust grasping of planar polygons, and provides experimental evidence of the value of the approach by a direct comparison between certified grasping and force-closure grasping.

%

The formulation we provide in this paper for each of the proposed certificates presents limitations--and opportunities for future work--which we detail in Sec.~\ref{sec:discussion}. Most notably, the presented formulation is purely geometrical, and does not take into account friction uncertainty, which can yield undesired behaviors between fingers and object such as jamming and wedging.



\section{Background}

This work inherits ideas from three main sources related to grasping and robustness:

\myparagraph{Sensorless Grasping.} Stemming from the foundational works by Mason and Erdmann on sensorless manipulation \cite{erdmann1988exploration}, and by Goldberg on sequences of squeezing grasps \cite{goldberg1993orienting}, this line of work aims to find grasping strategies that reliably bring an object to a known configuration, despite initial uncertainty in the object pose. In \cite{goldberg1993orienting}, Goldberg proposes an algorithm to find squeezing grasps that can reorient any convex polygon. This can be seen as a particular case of conformant path planning \cite{lozano1984automatic,erdmann1988exploration}, which synthesizes motions that drive a robot from an initially uncertain pose towards a goal, possibly under uncertain dynamics. This paper maintains the spirit of these works and studies the case of general point-based manipulators, and general planar polygonal objects. 

\myparagraph{From caging to grasping.} One way to constrain the object configuration to an invariant set is to cage it \cite{rimon1996caging}. While not all cages lead to a grasp \cite{rodriguez2012caging}, these always provide a certificate that the object is bounded to some compact set. More importantly, some cages are guaranteed to have a motion of the fingers that drives the cage into a grasp of the object. We are interested in synthesizing cages that lead to an unique grasp. 

\myparagraph{Computational models for caging.}  Many algorithms for cage synthesis have been studied since its introduction \cite{rimon1996caging}. The most relevant to this work is the optimization model in \cite{aceituno-cabezas2019icra}, which poses the caging condition in terms of convex-combinatorial constraints. We exploit the properties of this model to include requirements of convergence of the grasp process and observability of the final grasp. Caging has also been studied in the context of randomized planning \cite{varava2017herding,varavafree2018}, making no assumptions on shapes, and graph-search defined on contact-space \cite{allen2015robust,bunis2018equilateral}, with polynomial bounds in complexity.\vspace{6pt}

Beyond these three main sources, other works have also studied the role of uncertainty in grasping from a more practical perspective. Zhou et al \cite{zhou2017probabilistic} handle uncertainty by exploiting models for contact and sliding. Here, as in \cite{goldberg1993orienting}, we limit our analysis to the configuration space, without accounting for frictional interaction nor contact dynamics. In exchange, we are able to synthesize a grasping trajectory that drives a large set of initial configurations to a goal grasp for any planar polygonal object.

\subsection{Preliminaries and Notation}

We define an object $\mathcal{O}$, on a workspace $\mathcal{W} \subseteq \mathbb{R}^2$, as an union of $M$ convex polygons $\mathcal{O} = \bigcup_{i=1}^M \boldsymbol{P}_{i}$. The boundary of the object is described by the union of $L$ line segments $\partial \mathcal{O} = \bigcup_{j=1}^L \boldsymbol{L}_{j}$. The complement of the object is the region $\mathcal{W} \setminus \mathcal{O} = \bigcup_{k=1}^R \mathcal{R}_k$ consisting of $R$ convex polygonal regions $\mathcal{R}_k$. \vspace{6pt}

We denote the \textit{Configuration Space} of $\mathcal{O} \subseteq SE(2)$ at instant $t$ as $\mathcal{C}$. We refer to a plane of $\mathcal{C}$ with fixed orientation $\theta$ as a $\mathcal{C}-$slice, denoted $\mathcal{C}(\theta)$. We refer to an arrangement of point fingers as the manipulator $\mathcal{M}$. We assume $\mathcal{M}$ has $N$ point fingers with positions $\mathcal{M} = \lbrace \boldsymbol{p}_1, \hdots, \boldsymbol{p}_N \rbrace \in \mathcal{W}^N$. We refer to the set of configurations where the object penetrates a finger as $\mathcal{C}$-obstacles. Then, the free-space of the object $\mathcal{C}_{free}(\mathcal{O},t)$ corresponds to the space $\mathcal{C}$ not intersecting any of the $\mathcal{C}$-obstacles. \vspace{6pt}

At time $t$, an object configuration $\boldsymbol{q} = [q_x,q_y,q_\theta]^T$ is caged if $\boldsymbol{q}$ lies in a compact-connected component of $\mathcal{C}_{free}(\mathcal{O},t)$ (or invariant set), denoted as  $\mathcal{C}^{compact}_{free}(\mathcal{O},t)$. Given a compact-connected component $\mathcal{A} \subset \mathcal{C}$, we refer to its \textbf{limit orientations} $\theta_u,~\theta_l$ to the maximum and minimum of $\theta$ in $\mathcal{A}$.

We will describe in Section \ref{sec:caging}, how the caging condition can be transcribed as a set of convex-combinatorial constraints when the object is represented as an union convex polygons and the manipulator is a set of point-fingers.

\section{Problem description} 

\label{sec:overview}
The problem of interest for this paper is that of finding a grasping motion that is certified to succeed. Formally, we define this problem as:\vspace{6pt}

\textbf{Problem 1 (Certified Grasping)}:  {Given an object $\mathcal{O}$, a manipulator $\mathcal{M}$, $S$ samples of $\mathcal{C}-$slices, and a goal object configuration $\boldsymbol{q}$, find a manipulator trajectory $\rho_\mathcal{M} = \lbrace \mathcal{M}(t) \  | \ t \in \lbrace 1,\dots,{N_T}  \rbrace \rbrace$ and a set $Q_0 \subset \mathcal{C}(\mathcal{O})$, such that $\rho_\mathcal{M}$ will drive any configuration of the objlpect $\boldsymbol{\hat{q}} \in Q_0$ towards an observable grasp on $\boldsymbol{q}$.}\vspace{6pt}

This problem can be seen as a particular case of the general problem known as LMT \cite{lozano1984automatic}, and as a generalization of Goldberg's squeezing plans \cite{goldberg1993orienting} for non-convex objects and point-finger contacts. For an object on a plane without friction, a solution to this problem results from implementing the certificates described in the previous section as a three-step process (discretized as a manipulator trajectory of $N_T$ time-steps):
\begin{itemize}
    \item {Invariance:} The configuration of the object $\boldsymbol{q}$ lies in a compact-connected component of its free-space. We will impose this condition at $t = 1$ with a convex-combinatorial model of caging \cite{aceituno-cabezas2019icra}.
    
    \item {Convergence:} The manipulator path drives all configurations in the initial invariant set (cage) towards the goal $\boldsymbol{q}$. To meet this condition, once the object is caged ($t \in \lbrace 2,\dots,N_T \rbrace$), the manipulator follows a penetration-free path over which the compact-connected component contracts. Then, at the final time-step of the path, the $\mathcal{C}$-obstacles reduce $\mathcal{C}^{compact}_{free}(\mathcal{O},N_T)$ to a singleton $\lbrace \boldsymbol{q} \rbrace$.
    
    \item {Observability:} As a consequence of the fingers motion, the final contact configuration can recover the object pose at $\boldsymbol{q}$ through proprioceptive sensing. We call such a configuration an \emph{observable grasp} and is a condition solely required at the end of the path ($t = N_T$).
\end{itemize}
The satisfaction of these constraints would give a geometric certificate that any configuration of the object in the set $Q_0 = \mathcal{C}^{compact}_{free}(\mathcal{O},t = 1)$ will be driven towards and immobilized in the goal grasp. The following three sections provide a model for each of these three steps, which then will be combined into an optimization problem (\textbf{MIQP1}) for certified grasping of polygonal objects.



\begin{figure}[t]
    \centering
    \includegraphics[height=0.25\linewidth]{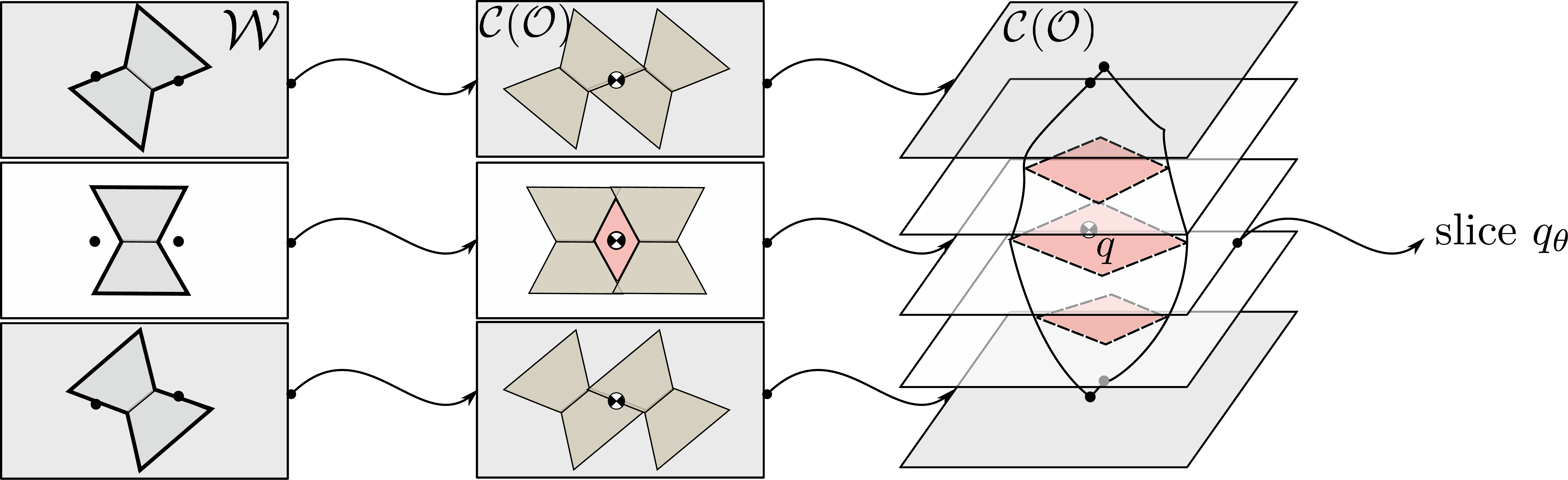}
    \caption{\textbf{Invariance Certificate.} Example of a cage in $\mathcal{W}$ (left), the $\mathcal{C}-$slices (center) and the configuration space $\mathcal{C}(\mathcal{O})$ (right). Note how the configuration $\boldsymbol{q}$ lies in a compact connected-component of the free-space (pink), bounded by two limit orientations (gray). Image adapted from \cite{aceituno-cabezas2019icra}.}
    \label{fig:cage_Ex}
\end{figure}

\section{Invariance Certificate} 
\label{sec:caging}

As explained above, one way to constrain an object to an invariant set is to cage it geometrically. Under the model presented in \cite{aceituno-cabezas2019icra}, the following are a set of sufficient conditions for invariance:

\begin{enumerate}
    \item The component $\mathcal{C}^{compact}_{free}(\mathcal{O},t)$ is bounded in the orientation coordinate by two limit orientations, otherwise it infinitely repeats along such axis with period $2 \pi$.
    \item At all $\mathcal{C}-$slices between the two limit orientations (when these exist) there is a loop of $\mathcal{C}-$obstacles enclosing a segment of free-space. All these loops must be connected, enclosing a component of free-space in between adjacent slices. At the slice with $q_\theta$, the loop must enclose $\boldsymbol{q}$ (as illustrated in the middle column of Fig. \ref{fig:cage_Ex}). 
    \item At the $\mathcal{C}-$slice of a limit orientation (if these exist) the free-space component enclosed by the loop has zero area. Thus, getting reduced to a line segment or a point.
\end{enumerate}

The union of these conditions define a net of constraints that enclose the configuration $q$, as illustrated in Fig. \ref{fig:cage_Ex}. Such conditions can be transcribed as a convex-combinatorial model composed of two sets of constraints, briefly described below, and explained in more detail in \cite{aceituno-cabezas2019icra}.

\subsection{Creating loops at each $\mathcal{C}$-slice}

\begin{figure}[t]
    \centering
    \includegraphics{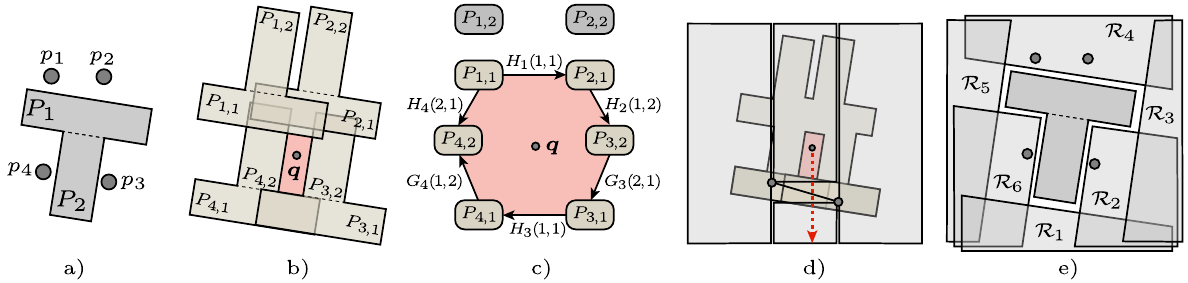}
  s  \caption{\textbf{Caging Model.} (a) Illustration of the cage of an object composed of two polygons ($M = 2$), caged with four fingers ($N = 4$) in a configuration space slice of constant orientation defined by six polygonal regions ($R = 6$), and with a boundary with eight edges ($L = 8$). (b) The model forms a polygonal loop at each slice of $\mathcal{C}(\mathcal{O},t)$, (c) defining a graph of polygonal intersections that enclose $\boldsymbol{q}$. (d) We test that the configuration $\boldsymbol{q}$ is enclosed by the loop by checking the red ray has an odd number of intersections with the loop. (e) Slightly exploded view of the (intersecting) polygonal regions that define the non-penetration space where the fingers can move.}
    \label{fig:f3}
\end{figure}

To construct a loop of $\mathcal{C}-$obstacles at each slice, we transcribe the problem as that of finding a closed directed graph within the intersections between polygonal obstacles. In such graph, each node represents a convex polygon of the decomposition of a $\mathcal{C}-$obstacle, while each edge imposes an intersection between polygons. We denote the polygon $i$ of $\mathcal{C}-$obstacle $n$ as $\boldsymbol{P}_{n,i}$. Including this condition in the model, at each time $t$, is done through the following constraints:

\myparagraph{Existence of a Loop.} This is encoded through two binary matrices: $\boldsymbol{H}_n \in \{0,1\}^{M \times M} \ \text{and} \ \boldsymbol{G}_n \in \{0,1\}^{M \times M}$. $\boldsymbol{H}_n$ encodes edges between $\mathcal{C}-$obstacle $n$ and $\mathcal{C}-$obstacle $n+1$, such that $\boldsymbol{H}_n(i,j) = 1 \Rightarrow \boldsymbol{P}_{n,i} \cap \boldsymbol{P}_{n+1,j} \neq \emptyset$. $\boldsymbol{G}_n$ encodes edges within $\mathcal{C}-$obstacle $n$, such that $\boldsymbol{G}_n(i,j) = 1 \Rightarrow \boldsymbol{P}_{n,i} \cap \boldsymbol{P}_{n,j} \neq \emptyset$. These matrices are constrained so that the resulting graph is closed and directed. We show an example of this loop and its graph in Fig. \ref{fig:f3} (b) and (c).

\myparagraph{Configuration Enclosing.} We include this condition by introducing a binary tensor $\mathbf{F} \in \{0,1\}^{N \times M \times 4}$, where $\mathbf{F}(i,j,k=1) = 1$ imposes a ray intersection with polygon $j$ at $\mathcal{C}-$obstacle $i$, while other values of $k$ assign the ray to the complement of the segment. The constraint needed to enclose $\boldsymbol{q}$ is to impose $\sum_{(i,j)} \mathbf{F}(i,j,k=1)$ to be odd. An illustration of this condition is shown in Fig. \ref{fig:f3} (d).

\myparagraph{Non-Penetration Constraints.} We impose this constraint by introducing a binary matrix $\boldsymbol{R} \in \{0,1\}^{N \times R}$. $\boldsymbol{R}(i,r) = 1$ assigns finger $i$ to region $r$ in $\mathcal{W} \setminus \mathcal{O}$, with $\sum_r \boldsymbol{R}(i,r) = 1, \forall i$. A visualization of this is shown in Fig. \ref{fig:f3} (e).\vspace{6pt}

Combining all of these constraints ensures the existence of a loop at each $\mathcal{C}-$slice and that $\boldsymbol{q}$ is enclosed by one of these loops.

\subsection{Constructing a cage from loops}

The next step is to impose that these constraints are only active for slices between two limit orientations (when these exist) while also enclosing a component of free-space between slices.

\myparagraph{Constraint Activation.} To determine which slices must contain a closed loop of $\mathcal{C}-$obstacles, we must first determine if the cage has limit orientations. To include this constraint, we introduce a binary vector $\Theta \in \{0,1\}^{S}$, where $\Theta(s) = 1$ imposes that a limit orientation must be reached before slice $s$, deactivating all loop constraints in such slice. In this context, \textit{before} means a greater or equal angle if the slice lies in the negative orientation half-space or a smaller or equal angle if it lies in the positive one.

\myparagraph{Limit Orientations.} A limit orientation occurs when the loop encloses a zero-area component, a condition defined by the contacts between the fingers and some translation of the object. To verify the existence of limit orientations, we define a binary matrix $\boldsymbol{T}_s \in \{0,1\}^{N \times L}$, such that $\boldsymbol{T}_s(i,l) = 1 \Rightarrow \boldsymbol{p}_i \in \boldsymbol{L}_l$ imposes that finger $i$ must be in contact with facet $l$ at slice $s$. Using this variable and labeling $\mathcal{L}_\mathcal{O}$ as the set of contact assignments that lead to a limit orientation, we impose $\boldsymbol{T}_s \in \mathcal{L}_\mathcal{O} \Rightarrow \Theta(s) = 1$.

\myparagraph{Continuous Boundary Variation.} In order for the $\mathcal{C}_{free}^{compact}(\mathcal{O},t)$ to be compact and connected, the loops created at the $\mathcal{C}-$slices must also enclose a segment of free-space between the slices. \cite{aceituno-cabezas2019icra} shows that a sufficient condition for this is to have the boundary of such loops to variate continuously unto the boundary of the loop in the adjacent $\mathcal{C}-$slices. A set of constraints for this condition are integrated as part of the model.

Satisfying these conditions ensures that the configuration $\boldsymbol{q}$ is enclosed by a compact-connected component of free-space. For more details on implementation and proofs on the correctness of these conditions, the reader is referred to \cite{aceituno-cabezas2019icra}.

\section{Convergence Certificate}
\label{sec:convergence}

Given an initial cage, the convergence certificate is satisfied if the process drives a set of bounded configurations towards the goal $\boldsymbol{q}$. The main insight that allows us to integrate this stage in the framework comes from the following remark:\vspace{6pt}

\textbf{Remark 1:} Given an object $\mathcal{O}$, at some time-step $t$, with a configuration $\boldsymbol{q}$ enclosed in a compact connected component of free-space $\boldsymbol{q} \in \mathcal{C}_{free}^{compact}(\mathcal{O},t)$ and bounded between limit orientations $\theta_{l}(t)$ and $\theta_{u}(t)$, any collision-free manipulator path $\rho_\mathcal{M}$ where $\mathcal{M}(N_T)$ immobilizes $\mathcal{O}$ at $\boldsymbol{q}$ and satisfies $\frac{d}{dt} \left(\theta_{u}(t) - \theta_{l}(t)\right ) < 0$ will drive any configuration $\boldsymbol{\hat{q}} \in \mathcal{C}_{free}^{compact}(\mathcal{O},t)$ towards $\boldsymbol{q}$. \vspace{6pt}

\begin{figure}[t]
    \centering
    \includegraphics[height=0.25\linewidth]{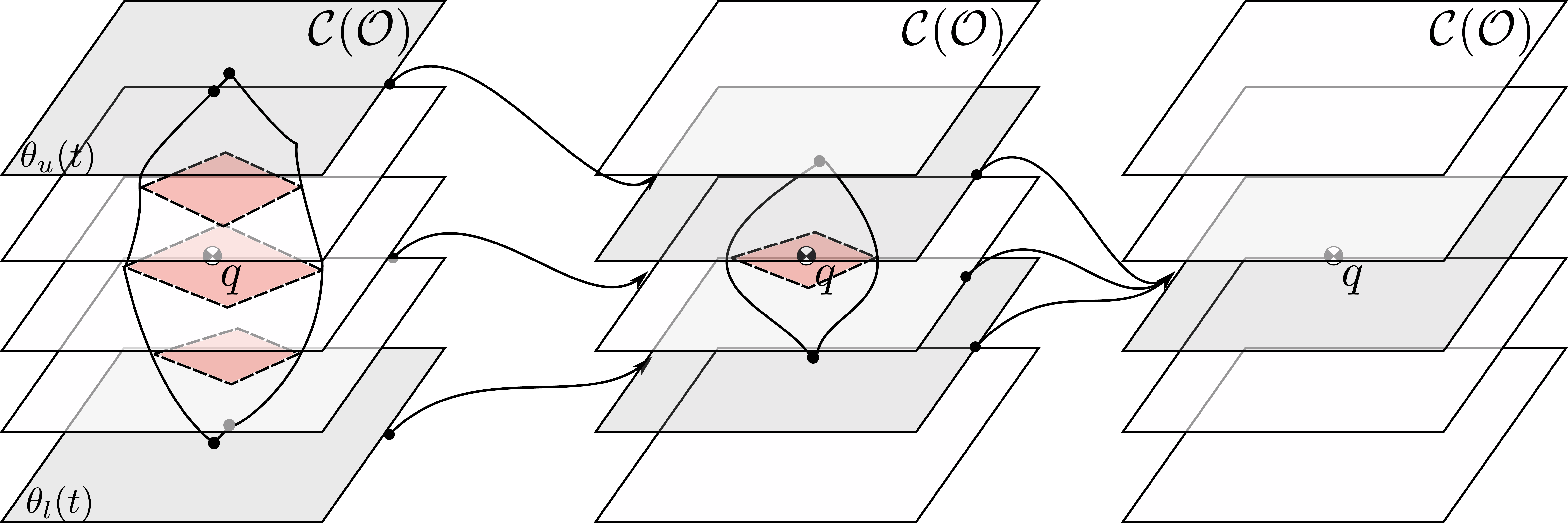}
     \caption{\textbf{Convergence Certificate.} This condition is trivially satisfied when the range of limit orientations (gray) decreases, converging at $t = N_T$.}
    \label{fig:f5}
\end{figure}

The conditions specified in Remark 1, shown in Fig. \ref{fig:f5}, are sufficient but might not be necessary. However, these allow us to optimize a manipulator path that satisfies the convergence certificate. This also allows us to characterize the set of initial configuration that will certifiably converge to $\boldsymbol{q}$. Hence, by relying on the model described in the previous section, we derive a linear model to certify convergence as detailed below.

\subsection{Certificate Model}

In order for the conditions detailed in remark 1 to hold, we require that:

\begin{enumerate}
    \item The object configuration must lie in a cage at all times. 
    \item The separation between \textit{limit orientations} must decrease monotonically between time-steps, until they converge at $t = N_T$.
    \item The cage at $t = N_T$ must only enclose the goal configuration $\boldsymbol{q}$.
\end{enumerate}
Algebraically, the conditions to impose a cage at each time-step are posed as:
\begin{eqnarray}
    \label{eq:cage}
    \begin{cases}
    \sum_{s} \Theta_s(t) = 2\\
    \theta_s(t) \in [\theta_l(t),\theta_u(t)] \Rightarrow \text{(loop existence)} |_{(s,t)} \\
    \end{cases}
\end{eqnarray}
for all $t \in \lbrace 1,\dots,N_T \rbrace$. Then, in order to ensure that the cage does not break between time-steps, we introduce the following constraint at each slice:
\begin{equation}
    \label{eq:non_break}
    \boldsymbol{H}_n(i,j)|_{t=k} \Rightarrow \exists ~ r_t \in \mathbb{R}^2 \ s.t. \ r_t \in \boldsymbol{P}_{i,n,k+1} \cap \boldsymbol{P}_{j,n+1,k+1}
\end{equation}
Note that this condition is sufficient and necessary, as the intersection occurs between convex polygons and the path is linearly interpolated. Because of this, we introduce the following remark:\vspace{6pt}

\textbf{Remark 2:} Since all initial configurations of the object are caged at $t = 1$ and the enclosing loop does not change between adjacent time-steps, the conditions for $\boldsymbol{q} \in \mathcal{C}_{free}^{compact}(\mathcal{O},t)$ are trivially satisfied for all $t>1$.\vspace{6pt}

Furthermore, for the final cage to fully immobilize the object, we require that there exist two similar \textit{limit orientations} at $t = N_T - 1$ which have the same facet assignment matrix, also enforced for $t =  N_T$ (when limit orientations converge). Note that this reduces $\mathcal{C}_{free}^{compact}(\mathcal{O},t = N_T)$ to a singleton. Algebraically, this constraint is added as:
\begin{eqnarray}
\begin{cases}
    \label{eq:limcont1} \boldsymbol{T}_{u}(N_T) = \boldsymbol{T}_{u}(N_T-1) = \boldsymbol{T}_{l}(N_T-1) \in \mathcal{L}_{\mathcal{O}} \\  |\theta_{u}(N_T-1) - \theta_{l}(N_T-1)| \approx 0 \label{eq:limcont2}
\end{cases}
\end{eqnarray}
Finally, the limit orientations converge gradually under the constraint:
\begin{eqnarray}
\begin{cases}
    \label{eq:cont1}\theta_{u}(t+1) < \theta_{u}(t) \\
    \label{eq:cont2}\theta_{l}(t) < \theta_{l}(t+1)
\end{cases}
\end{eqnarray}
This, along with the caging model, certifies that the grasp will always succeed within a set of certified initial configurations $Q_0$. 

\myparagraph{Constraining $Q_0$.} Constraining that $\mathcal{C}_{free}^{compact}(\mathcal{O})$ contains an arbitrary set of initial conditions $Q_0$ cannot be integrated in general within this convex-combinatorial model. However, we can use an inner box approximation of $Q_0$ in the form: $Q_0 = \lbrace q \in \mathcal{C}(\mathcal{O}) \ | \ q \in [x_1,x_2] \times [y_1,y_2] \times [\theta_1,\theta_2] \rbrace$ by adding the following constraints: 
\begin{eqnarray}
    \label{eq:incond}
        \theta_s \in [\theta_1,\theta_2] \Rightarrow (\text{configuration enclosing}), \ \forall (x,y) \in [x_1,x_2] \times [y_1,y_2]
\end{eqnarray}
In this case, robust optimization \cite{ben2009robust} would be used in each $\mathcal{C}-$slice to ensure that all points in $[x_1,x_2] \times [y_1,y_2]$ are enclosed by the cage.

\section{Observability Certificate}
\label{sec:observability}

Once a planned grasp process has been executed, we can also certify the immobilization at the goal configuration if the grasp is \textit{observable}, i.e.~such that we can retrieve the object pose from sensor readings. In this section, we present a definition of grasp observability and derive sufficient constraints for a grasp to be locally observable under proprioceptive sensing (e.g. joint encoders). In practice, this adds an extra constraint to the type of end grasp that we are interested in.

\subsection{Definitions}

Given a vector of $n_r$ sensor readings $\boldsymbol{s} \in \mathbb{R}^{n_r}$, we define:\vspace{6pt}

\textbf{Definition 1} (Sensor Model):  Given a final grasp $G$ achieved by a manipulator configuration $\mathcal{M}(N_T)$, we define a sensor model $F_G$ as a mapping from object configurations to sensor readings:
$$\begin{array}{llll}
F_G: ~ & \mathcal{C}(\mathcal{O}) & \longrightarrow & \mathbb{R}^{n_r} \\
& \boldsymbol{\hat{q}} & \longmapsto & \boldsymbol{s} = (s_1, ..., s_{n_r}) = F_G(\boldsymbol{\hat{q}}).
\end{array}$$

\textbf{Definition 2} (Grasp Observability):  Given a grasp $G$, a sensor model $F_G$ and a final object configuration $\boldsymbol{q}$, we will say that $G$ is observable if and only if $F_G$ is locally invertible around $\boldsymbol{q}$.\vspace{6pt}

\begin{wrapfigure}{R}{0.6\textwidth}
    \centering
    \vspace{-24pt}
    \includegraphics[height=0.38\linewidth]{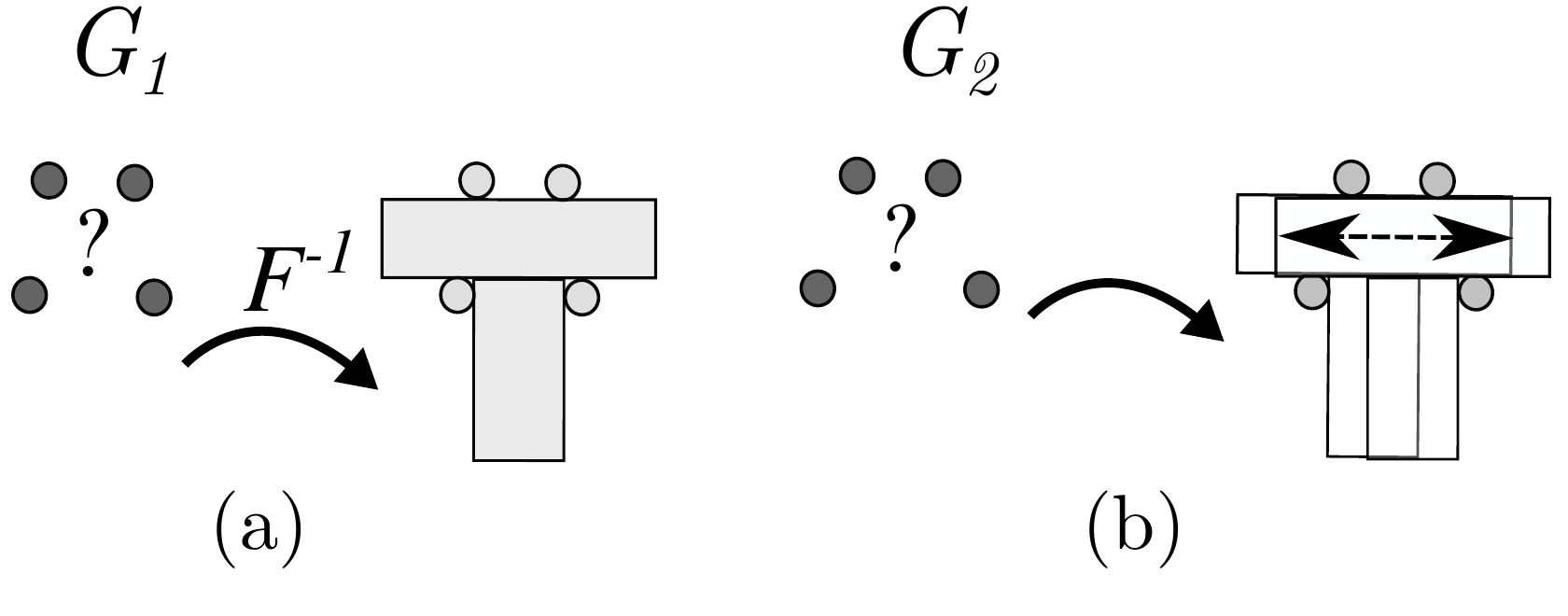}
    \caption{\textbf{Observability Certificate.} (a) An observable grasp $G_1$. (b) A non-observable grasp $G_2$, the object can slide between the fingers.}
    \vspace{-24pt}
    \label{fig:fobs}
\end{wrapfigure}

\textbf{Remark 3}: If $n_r \geq 3$ and the sensor model $F_G$ satisfies that its Jacobian $JF_G(\boldsymbol{q}) \in \mathbb{R}^{n_r \times 3}$ is full rank, then the grasp $G$ is observable and only 3 sensor readings are necessary for observability.\vspace{6pt}

Fig. \ref{fig:fobs} shows an example of grasp observability. In general, $F_G$ can be hard to define in closed form, as it depends on the object and manipulator geometries. Hence, we restrict our analysis to first order effects \cite{rimon1996force}.

\myparagraph{Proprioceptive Sensor Model:} In order to give an intuitive notion of a sensor reading, we characterize a sensor model for point-contact sensing to first order effects. Intuitively, for an object in contact, this model reports local changes based on a gap function at each contact point, $\psi_i(\boldsymbol{\bar{q}}, \boldsymbol{p}_i)$, as it is commonly used to formalize the study of grasp stability \cite{prattichizzo2016grasping}. More concretely, sensor readings should only report changes in the object pose that imply decrements of the gap (causing penetration), ignoring changes that preserve or break contact (no applied force). 
Therefore, we characterize a sensor reading with the result of applying the sensor model Jacobian to an infinitesimal object configuration variation $d\boldsymbol{q}$ from $\boldsymbol{q}$:
$$
JF_G(\boldsymbol{q}) = \left( \begin{matrix} \frac{ds_1}{d \boldsymbol{q}}(\boldsymbol{q}) \\ \vdots \\ \frac{ds_{n_r}}{d \boldsymbol{q}}(\boldsymbol{q}) \end{matrix} \right), \hspace{5pt} \frac{ds_i}{d \boldsymbol{q}}(\boldsymbol{q}) ~ d\boldsymbol{q} = 
\begin{cases}
k_i~\dfrac{d \psi_i}{d \boldsymbol{q}}(\boldsymbol{q},\boldsymbol{p}_i)~d\boldsymbol{q},  & \dfrac{d \psi_i}{d \boldsymbol{q}}(\boldsymbol{q},\boldsymbol{p}_i)~d\boldsymbol{q} < 0, \\[9pt]
0, & \dfrac{d \psi_i}{d \boldsymbol{q}}(\boldsymbol{q},\boldsymbol{p}_i)~d\boldsymbol{q} \geq 0,
\end{cases}
$$
where $k_i$ is a real non-zero constant. \vspace{6pt}

The first-order behavior of the proprioceptive sensor model above highlights a relation between observability and first-order form closure. As a result of Remark 3, we will consider only three sensor readings, $n_r = 3$. 

\vspace{6pt}\textbf{Remark 4}: Given a grasp $G$ of an object in its final configuration $\boldsymbol{q}$, first-order form closure is equivalent to have the matrix $JF_G(\boldsymbol{q})$ be of full rank, where $F_G$ is the proprioceptive sensor model defined above.

\begin{proof}
Note that having full rankness of $JF_G(\boldsymbol{q}) \in \mathbb{R}^{3 \times 3}$ is equivalent to:
$$
\left[ JF_G(\boldsymbol{q}) ~ d\boldsymbol{q} = \boldsymbol{0} \Rightarrow d\boldsymbol{q} = \boldsymbol{0} \right] ~\Leftrightarrow~ \left[
\forall~i, ~~ \frac{ds_i}{d \boldsymbol{q}}(\boldsymbol{q}) ~ d\boldsymbol{q} = 0 \Rightarrow d\boldsymbol{q} = \boldsymbol{0} \right].$$
As a result of the first-order behavior of our virtual sensor model, we have
$$
\frac{ds_i}{d \boldsymbol{q}}(\boldsymbol{q}) ~ d\boldsymbol{q} = 0 ~\Leftrightarrow~ \dfrac{d \psi_i}{d \boldsymbol{q}}(\boldsymbol{q},\boldsymbol{p}_i)~d\boldsymbol{q} \geq 0,
$$
where the implication from right to left is by definition and from left to right is a consequence of $k_i \neq 0$. Therefore, 
$$\left[
\forall~i, ~~ \frac{ds_i}{d \boldsymbol{q}}(\boldsymbol{q}) ~ d\boldsymbol{q} = 0 \Rightarrow d\boldsymbol{q} = \boldsymbol{0} \right] \Leftrightarrow \left[ \forall~i, ~~ \frac{d \psi_i}{d \boldsymbol{q}}(\boldsymbol{q}, \boldsymbol{p}_i) ~ d\boldsymbol{q} \geq 0 \Rightarrow d\boldsymbol{q} = 0 \right],$$
that is precisely a characterization of first-order form closure \cite{prattichizzo2016grasping}. Consequently, first-order form closure is equivalent to full rankness of $JF_G(\boldsymbol{q})$, when considering $F_G$ as the proprioceptive sensor model.
\end{proof}

\textbf{Corollary 1}: Given a grasp $G$ and the proprioceptive sensor model $F_G$, first-order form closure implies grasp observability.




\subsection{Certificate Model}

\begin{figure}[t]
    \centering
    \begin{minipage}{0.48\linewidth}
        \centering
        \vspace{15pt}
        \includegraphics[height=0.32\linewidth]{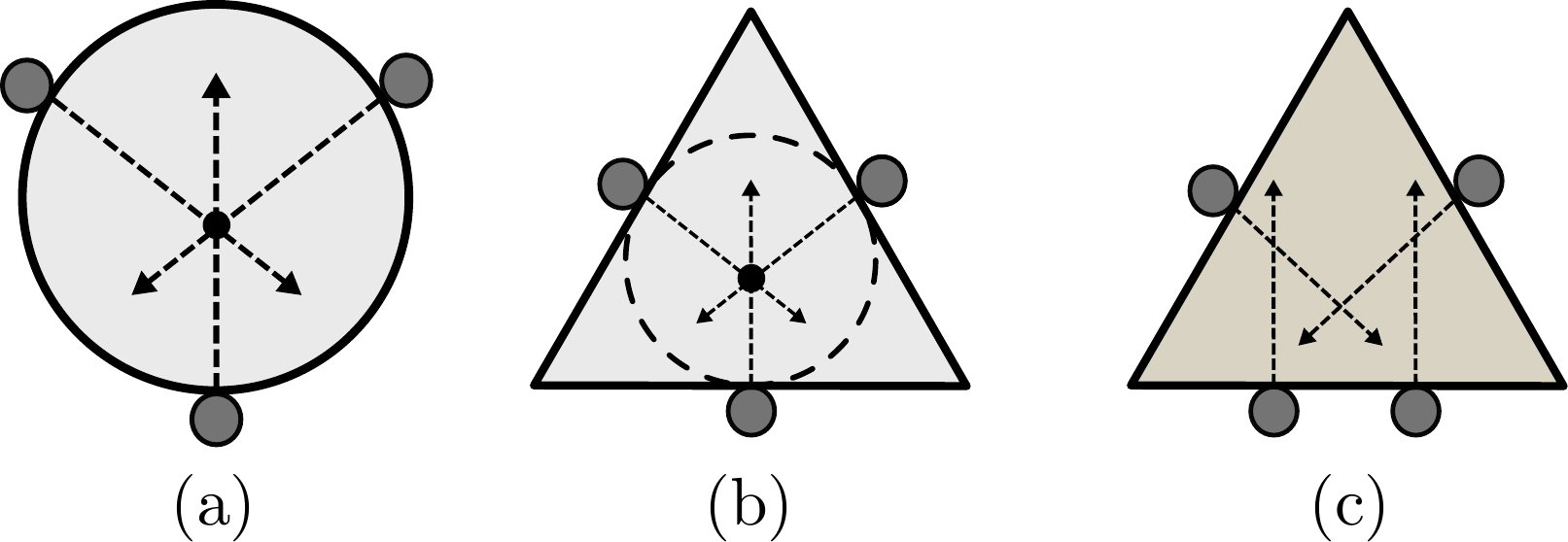}
        \vspace{6pt}
        \caption{Examples of proprioceptive observability conditions: (a) Not observable, (b) First-Order Not-Observable, and (c) First-Order Observable.}
        \label{fig:f9}
    \end{minipage}
    ~
    \begin{minipage}{0.48\linewidth}
        \centering
        \includegraphics[height=0.39\linewidth]{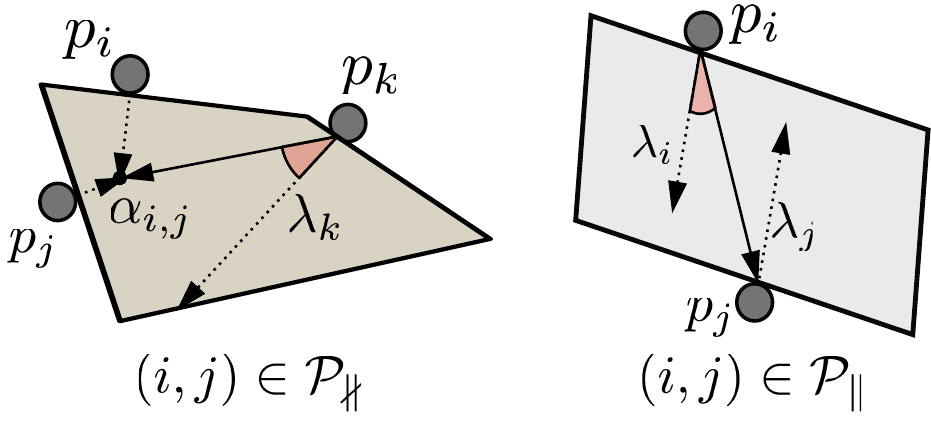}
        \caption{Convex combinatorial conditions for non-coincidence in the case of non-parallel (left) and parallel (right) facet assignments.}
        \label{fig:f10}
    \end{minipage}
    \vspace{-12pt}
\end{figure}
Given the relation between form-closure and observability that we derived above, a planar grasp is first-order observable if there are $4$ unilateral contact constraints on the object \cite{rimon1996force}. This is satisfied if the following conditions hold:
\begin{enumerate}
    \item The object configuration must lie in a singleton of $\mathcal{C}_{free}(\mathcal{O},t)$. This condition is already implied by (CT3).  
    \item There must exist no point of coincidence between all the contact normals. This is required because otherwise, to first-order, the object would be free to rotate infinitesimally around the point of concurrency of the contact normals \cite{rimon1996force}.
\end{enumerate}

 Fig. \ref{fig:f9} shows examples. These are convex-combinatorial constraints on the facet-assignment matrix $\boldsymbol{T}_s$ and manipulator configuration $\mathcal{M}({N_T})$. Algebraically, the non-coincidence condition can be expressed as:
$$
\bigcap_{i}{\boldsymbol{p}_{i}(N_T) + \langle \lambda_i \rangle} = \emptyset
$$
We note that there are two scenarios for every pair of fingers: 1) Intersecting normals correspond to non-parallel facets and have a single intersection point, and 2) Normal vectors are parallel and thus have infinite intersection points or none. Therefore, if we define the following sets:
\begin{itemize}
    \item $\mathcal{P} = \{(i,j) \in N^2 \ | \ i > j \}$ is the set of all different pairs of facet-assignments.
    \item $\mathcal{P}_{\parallel} = \{(i,j) \in \mathcal{P} \ | \ \lambda_i \times \lambda_j = 0 \}$ is the set of pairs of facet-assignments with parallel normals.
    \item $\mathcal{P}_{\nparallel} = \{(i,j) \in \mathcal{P} \ | \ \lambda_i \times \lambda_j \neq 0 \}$ is the set of pairs of facet-assignments with nonparallel normals.
\end{itemize}
where $\times$ is the ordinary cross-product. Then, we can introduce the binary matrix $\boldsymbol{M} = (\boldsymbol{M}_{i,j})_{(i,j) \in \mathcal{P}} \in \{0,1\}^{|\mathcal{P}|_c}$, where $| \mathcal{P} |$ is the cardinality of $\mathcal{P}$, reducing the problem to the following set of convex-combinatorial conditions:
\begin{eqnarray} 
    \label{obs_1} \boldsymbol{M}_{(i,j) \in \mathcal{P}_{\nparallel}} \Rightarrow \sum_{k = 1}^N | (\alpha_{i,j} - \boldsymbol{p}_{k}(N_T)) \times \lambda_k | > 0
    \\ 
    \boldsymbol{M}_{(i,j) \in \mathcal{P}_{\parallel}} \Rightarrow | (\boldsymbol{p}_{i}(N_T) - \boldsymbol{p}_{j}(N_T)) \times \lambda_i | > 0
    \\ 
    \label{obs_i} \sum_{(i,j) \in \mathcal{P}} \boldsymbol{M}_{i,j} \geq 1, \label{obs_f}
\end{eqnarray}

where $\alpha_{i,j}$ is the intersection point between the lines defined by the normal vectors starting at $\boldsymbol{p}_{i}(N_T)$ and $\boldsymbol{p}_{j}(N_T)$. These conditions guarantee that at least one pair of normals is non-coincident to the rest, providing observability as shown in Fig. \ref{fig:f10}. Here, we include absolute value function through slack variables and big-M formulation \cite{floudas1995nonlinear}.

\section{Application to Sensorless Grasping} 
\label{sec:results}

This section describes an optimization problem for grasping of planar objects with bounded uncertainty. For this, we formulate a Mixed-Integer program (MIP) using the constraints described in sections 4, 5 and 6. We validate this approach on different polygonal objects, both with experiments and simulations. All the computations are done in MATLAB R2018b on a MacBook Pro computer with Intel Core i9 2.9 GHz processor. All optimization problems are solved with Gurobi 8.0 \cite{gurobi}.

\subsection{Mixed-Integer Programming Formulation}

We propose a formulation which receives as inputs the description of the polygonal object $\mathcal{O}$ and the manipulator $\mathcal{M}$. We incorporate the conditions described through the paper as constraints and add a quadratic cost term on acceleration to smooth the trajectory, resulting in problem \textbf{MIQP1}.

\begin{equation}\nonumber
\mathbf{MIQP1:} \ \underset{\substack{\mathcal{M}(t)}}{\text{\text{min}}} \ \ \int \sum_{i = 1}^{N} \left|\left| \frac{d^2 \boldsymbol{p}_i(t)}{d t^2} \right|\right|^2 dt
\end{equation}
subject to:
\begin{enumerate}
	\item  For $t = 1$ to $t = N_T$:
    \begin{itemize}
        \item Caging (CT1)-(CT2).
        \item Convergence Certificate (CT4)-(CT5).
    \end{itemize}
   \item $(t = 1)$ Invariance certificate \eqref{eq:incond}.
   \item $(t = N_T)$ First-Order Grasp Observability \eqref{obs_1}-\eqref{obs_f}.
\end{enumerate} 


\subsection{Simulated Experiments}
\begin{figure}[b]
    \centering
    \includegraphics[width=0.99\linewidth]{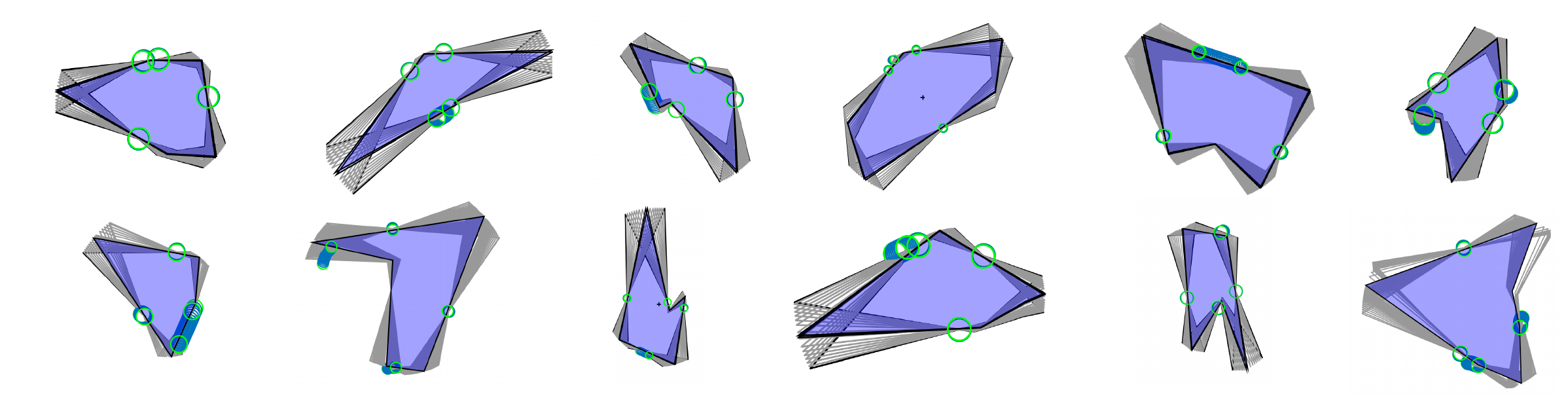}
    \caption{\textbf{Simulation results.} 12 random polygons are grasped with trajectories generated with our model. In each case, a set of random initial configurations certified by our model (shown in gray) are driven towards a goal grasp (purple) by using the same trajectory (blue).}
    \label{fig:f12}
\end{figure}
We generate a set of 12 random polygons and optimize a trajectory for each using \textbf{MIQP1}. Then, we perform simulations for a set of over 100 different initial conditions, using the open planar manipulation simulator in \cite{zhou2018convex}. We initialize the plan with limit orientations between $-15^\circ$ and $15^\circ$, centered around $\boldsymbol{q} = 0$. This limits certification for the configurations with $\theta \in [-15^\circ,15^\circ]$, with no hard guarantees on translational uncertainty.\vspace{6pt}

In order to generate random polygons with interesting properties, we rely on the heuristics presented in \cite{auer1996rpg}, which specify parameters such as irregularity and referential radius. We implement this code in MATLAB and generate the 12 polygons of Fig. \ref{fig:f12}, with 4 to 6 facets. We segment each object with Delaunay triangulation \cite{fortune1995voronoi} and determine $\mathcal{W} \setminus \mathcal{O}$ with hueristics valid for simple enough shapes. Is worth noting that algorithms other than Delaunay triangulation might be able to find a decomposition with a small number of convex polygons \cite{lien2006approximate}.\vspace{6pt}

For each initial condition, we execute the trajectory with 4 free disc-shaped fingers. Results for 12 of the random object are reported in Fig. \ref{fig:f12}. Using the same manipulator trajectory, a set of different initial poses (marked in gray) are driven towards $\boldsymbol{q}$ (blue). For all the objects, a trajectory was successfully found in 25 to 45 seconds. However, the time required to find the optimal trajectory ranged from several seconds to around two minutes, depending on the number of integer variables of the problem. We note that fixing limit orientations usually allows for little translational uncertainty, suggesting the need for (CT5) in the general case. \vspace{-12pt}

\subsection{Real Robot Experiments}

We demonstrate trajectories generated on four different planar objects in a real experimental set-up with a two-armed robot. We optimize trajectories for each of the objects in Fig. \ref{fig:f12} and use simulations to determine $Q_0$. Each trajectory is designed with $N_T = 5$ time-steps and initial limit orientations between $-22.5^\circ$ and $22.5^\circ$. We perform 10 experiments on each object, initializing them at random initial configurations within the invariant set $Q_0$.\vspace{6pt}
\begin{figure}[t]
    \centering
    \includegraphics[width=0.9\linewidth]{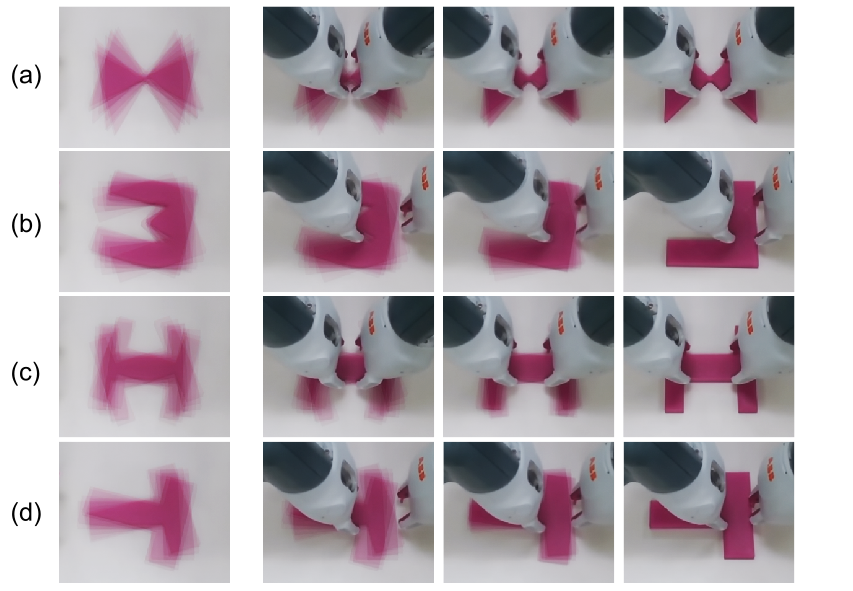}
    \caption{\textbf{Experimental results.} Each row shows snapshots from execution of the resulting grasping trajectories for 4 objects, overlaying 10 experiments with initial pose uncertainty (first frame) moving towards a single goal configuration (last frame). Our certification allows for significant rotational uncertainty in the initial object configuration, always converging to the same goal.}
    \vspace{-12pt}
    \label{fig:f11}
\end{figure}

Our robotic platform is an ABB YuMi$^{\tiny{\text{\textregistered}}}$ (IRB-14000) robot, which has two 7 DOF arms with parallel jaw grippers. We work with a Robot Operating System (ROS) setup and an Intel RealSense D415 RGB-D camera calibrated with AprilTag 2 scanning, which we use to place the object within the reach of the robot, and within the invariant set $Q_0$. Additional constraints are added to \textbf{MIQP1} to account for the kinematics of the manipulator. The end-effectors of YuMi$^{\tiny{\textregistered}}$ are modified to have thin cylindrical fingers. To showcase the robustness of this approach, all experiments are run open-loop.\vspace{6pt}

Fig. \ref{fig:f11} shows resulting trajectories for 10 different initial conditions of the four objects. Depending on the shape, the resulting trajectories vary from stretching -- (a) and (b) -- to squeezing (d), and a combination of both (c). In all cases, we are able to handle significant uncertainty in the orientation axis, and varying translational uncertainty (from millimeters to a few centimeters). Videos on the experiments for each of the objects are shown in the supplementary material. 

\subsection{Comparison with pure force-closure grasping}

A natural question is how accounting for certification compares to a naive reaching strategy. In order to provide a quantitative answer to this question, we compare our approach to a naive grasping plan which optimizes some criteria of grasp quality, as commonly done in grasp planning algorithms. We design this naive motion by searching for a force close grasp \cite{prattichizzo2016grasping} and approaching each contact with a trajectory perpendicularly to the goal facet, starting all fingers with the same separation. \vspace{6pt}

\begin{figure}[b]
    \vspace{-12pt}
    \centering
        \includegraphics{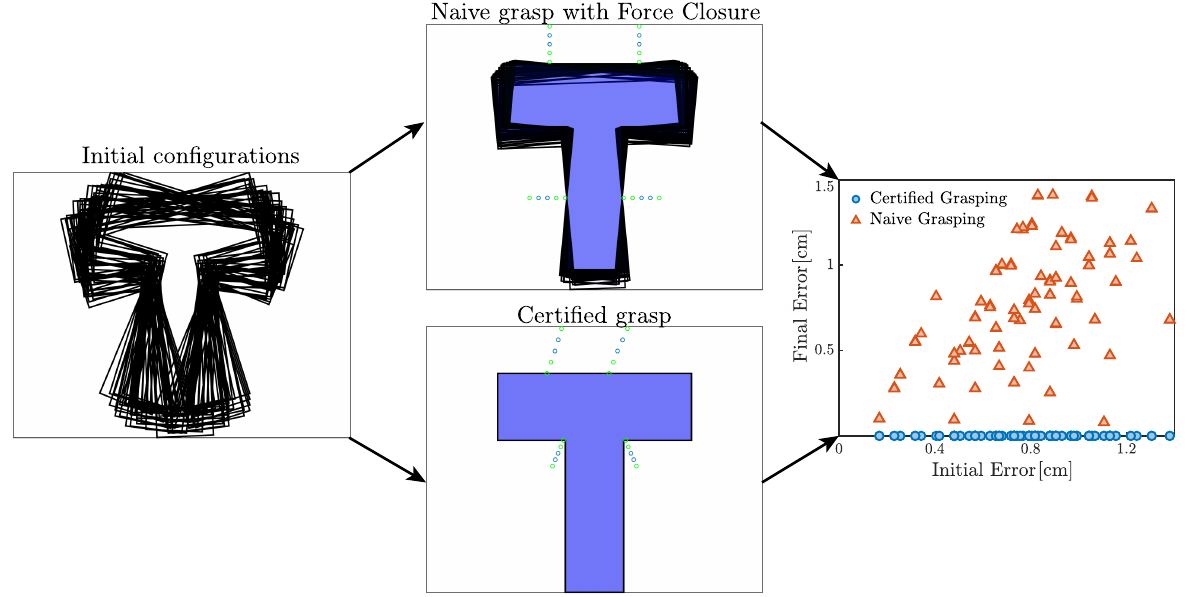}
    \caption{\textbf{Certified Grasping vs. Force-Closure Grasping.} We simulate grasps over an object with noisy initial configuration (left). A traditional grasping strategy that maximizes force closure (top-center) fails to handle uncertainty, resulting in significant error in the final pose of several simulations. In contrast, certified grasping (bottom-center) drives the object to its goal pose, always converging to the same configuration. By comparing the $L_1$ error on the final pose (right), we obtain that certified grasping is orders of magnitude more accurate than a naive policy.}
    \label{fig:f13}
\end{figure}

We simulate both strategies to grasp a T-shaped object from 100 different initial conditions. Certified grasping always drives the object to the goal configuration with proprioceptive observability. We measure the $L_1$ distance to desired object pose, which we call error, after each grasping strategy is executed and report our results in Fig. \ref{fig:f13}. As can be seen in many of these simulations, the naive force-closure grasp does not drive the object towards the goal nor does it provide observability.

\section{Discussion} 
\label{sec:discussion}

In this paper we study certified grasping of planar objects under bounded pose uncertainty. To do this, we extend grasp analysis to include the reaching motion towards the final arrangement of contacts. Under this perspective, we propose three certificates of grasp success: 1) Invariance within an initial set of configurations of the object, 2) convergence from the initial set to a goal grasp, and 3) observability of the final grasp. 

For each of the these certificates, we derive a mathematical model, which can be expressed with convex-combinatorial constraints, and demonstrate their application to synthesize robust sensorless grasps of polygonal objects. We validate these models in simulation and with real robot experiments, showcasing the value of the approach by a direct comparison with force closure grasping.


\myparagraph{Limitations.}
The first limitation of this work comes from restricting the analysis to the configuration space of the object. This neglects frictional interaction between the fingers and the object, which could lead to unaccounted stable configurations such as jamming or wedging. Accounting for the role of friction, characterizing undesired scenarios such as in~\cite{haas2018passive}, would allow this framework to provide certification over a larger range of dynamic settings. The second limitation comes from the first-order proprioceptive analysis of observability. Including second-order effects such as curvature of the object~\cite{rimon1996force} as well as accounting for more discriminative sensor models that provide shape, texture, or force information~\cite{donlon2018gelslim}, could certify success without requiring form-closure constraints.

\myparagraph{Future Work.}
Given the versatility of convex-combinatorial optimization models, we believe that this approach can be extended to the design of finger phalanges with complex shapes beyond finger points~\cite{rodriguez2013effector}. This would allow to certifiably grasp specific objects within a larger set of initial conditions and with a lower number of fingers. Additionally, we are interested in extending this model to invariance sets that are not purely geometrical, for example by considering energy bounds~\cite{mahler2016energy} or other type of dynamic constraints on object mobility. 

\bibliographystyle{plainnat}
{\footnotesize \bibliography{references}}

\end{document}